\documentclass[10pt,twocolumn,letterpaper]{article}

\usepackage{cvpr}              

\definecolor{cvprblue}{rgb}{0.21,0.49,0.74}
\usepackage[pagebackref,breaklinks,colorlinks,citecolor=cvprblue]{hyperref}
\usepackage{colortbl}  
\usepackage[dvipsnames]{xcolor}
\usepackage{comment}
\usepackage{multirow}
\usepackage{bm}  
\usepackage{fancyhdr}
\usepackage{cuted}
\def\confName{CVPR}
\def\confYear{2025}

\newcommand{\paravspace}{\vspace{-10pt}}

\title{\texttt{CogACT}: A Foundational Vision-Language-Action  Model for Synergizing Cognition and Action in Robotic Manipulation}

\makeatletter
\renewcommand*{\@fnsymbol}[1]{\ensuremath{\ifcase#1\or *\or \dagger\or \ddagger\or
		\mathsection\or \mathparagraph\or \|\or **\or \dagger\dagger
		\or \ddagger\ddagger \else\@ctrerr\fi}}
\makeatother

\author{Qixiu Li$^{1}$\thanks{Equal Contributions}\ \ \!\thanks{Interns at Microsoft Research}\,\, \  Yaobo Liang$^{2 *}$\thanks{Project Leads. Email: {\tt \{yalia,jiaoyan\}@microsoft.com} }  \ \ \ Zeyu Wang$^{1 * \dagger}$ \ Lin Luo$^{2}$ \ Xi Chen$^{2}$ \ Mozheng Liao$^{3 \dagger}$ \ Fangyun Wei$^{2}$ \\
Yu Deng$^{2}$ \ Sicheng Xu$^{2}$ \ Yizhong Zhang$^{2}$ \ Xiaofan Wang$^{4  \dagger}$ \ Bei Liu$^{2}$ \ Jianlong Fu$^{2}$ \ Jianmin Bao$^{2}$ \\ 
Dong Chen$^{2}$ \ Yuanchun Shi$^{1}$ \ Jiaolong Yang$^{2 \ddagger}$ \ \  Baining Guo$^{2}$\\
$^1${Tsinghua University} \quad $^2${Microsoft Research Asia} \quad $^3${USTC} \quad $^4${Institute of Microelectronics, CAS} \\
 \vspace{-58pt}
}

\begin{document}

\maketitle

\begin{strip}
	\centering
	\includegraphics[width=0.92\textwidth]{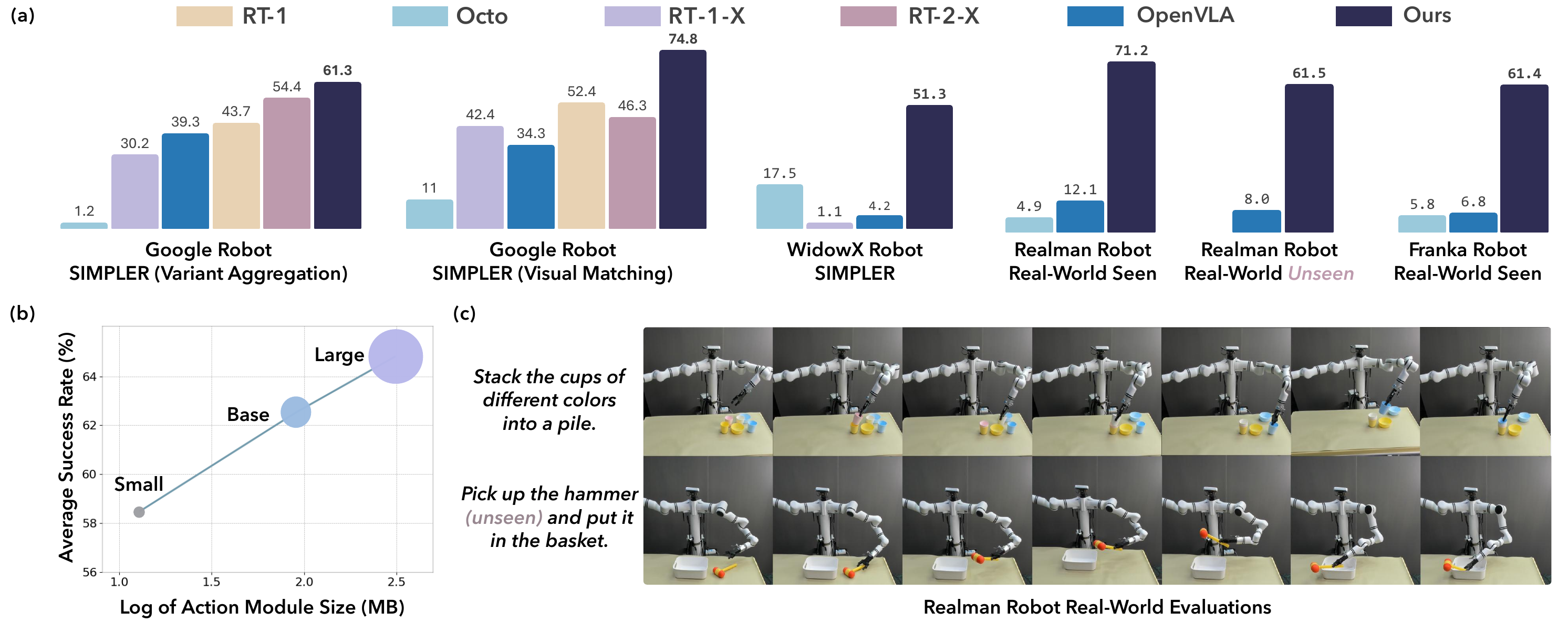}
    \vspace{-2pt}
	\captionof{figure}{
        (a) Success rate (\%) comparison of our model against RT-1~\cite{brohan2022rt1}, RT-1-X~\cite{o2023open_x_embodiment}, RT-2-X~\cite{o2023open_x_embodiment}, Octo~\cite{team2024octo}, and OpenVLA~\cite{kim2024openvla} across simulated benchmarks (first thee charts) and real-world evaluations (last three charts), using various robots: Google robot, WidowX robot, Realman robot, and Franka robot. All models are trained on the expansive Open X-Embodiment dataset~\cite{o2023open_x_embodiment} (expect for RT-1 that trains only on the Google Robot subset) and finetuned on a small amount of data for real robot experiments.
(b) Scaling behavior: averaged success rate on SIMPLER~\cite{li2024simpler} with respect to action module size.
(c) Examples of Realman Robot executing tasks involving sequentially stacking multiple cups and picking and placing unseen objects.
    }
    \label{fig:teaser}
	\vspace{-1pt}
\end{strip}

\begin{abstract}
The advancement of large Vision-Language-Action (VLA) models has significantly improved robotic manipulation in terms of language-guided task execution and  generalization to unseen scenarios. 
While existing VLAs adapted from pretrained large Vision-Language-Models (VLM) have demonstrated promising generalizability, their task performance is still unsatisfactory as indicated by the low tasks success rates in different environments. 
In this paper, we present a new advanced VLA architecture derived from VLM. Unlike previous works that directly repurpose VLM for action prediction by simple action quantization, we propose a componentized VLA architecture that has a specialized action module conditioned on VLM output. We systematically study the design of the action module and demonstrate the strong performance enhancement with diffusion action transformers for action sequence modeling, as well as their favorable scaling behaviors. We also conduct comprehensive experiments and ablation studies to evaluate the efficacy of our models with varied designs.
The evaluation on five robot embodiments in simulation and real work shows that our model not only significantly surpasses existing VLAs in task performance but also exhibits remarkable adaptation to new robots and generalization to unseen objects and backgrounds. It exceeds the average success rates of OpenVLA which has similar model size (7B) with ours by over 35\% in simulated evaluation and 55\% in real robot experiments. It also outperforms the large RT-2-X model (55B) by 18\% absolute success rates in simulation. Code and models can be found on our \href{https://cogact.github.io/}{project page}.

\vspace{-20pt}
\end{abstract}    
\section{Introduction}
\label{sec:intro}
In recent years, there has been a surge of interest in robotic control models  equipped with visual capabilities~\cite{brohan2022rt1,nair2023r3m,brohan2023rt2,stone2023open,shridhar2023perceiver,chi2023diffusion_policy,o2023open_x_embodiment,livision2024,team2024octo,wen2024tinyvla,wu2024unleashing,kim2024openvla}. Among them, the development of large-scale {V}ision-{L}anguage-{A}ction (VLA) models~\cite{brohan2023rt2,kim2024openvla,li2023vision} are particularly promising, which empowers robots to perform complex tasks guided by natural language instructions and potentially manage objects or environments that deviate from the training distribution. Additionally, they exhibit rapid adaptability to new tasks and embodiments through finetuning.

The notable generalization capability of large VLAs can be attributed to both their substantial model size and the potent Vision-Language-Models (VLM)~\cite{liu2023visual,chen2023pali,karamcheti2024prismatic} that serve as their foundation. These VLMs are typically pretrained on massive, Internet-scale image-text pairs, which play a crucial role in enhancing VLA generalization to novel objects and semantically diverse instructions~\cite{brohan2023rt2}.

Existing large VLAs often adapt VLMs for action prediction in simple ways, leading to several issues that hinder task performance. For instance, works like \cite{brohan2023rt2,kim2024openvla} directly quantize the continuous spectrum of robot actions into discrete bins in accordance to the next token prediction scheme of VLMs. However, such a simple quantization, unlike sophisticated tokenizers such as those designed for images~\cite{van2017neural,yu2023language} and audio~\cite{zeghidour2021soundstream,defossez2022high}, poses difficulties in action learning and limits action precision. \cite{li2023vision} introduces additional action heads, such as LSTMs, to transform VLM output into actions. The shift to a regression-based learning scheme, however, overlooks the probabilistic and multimodal\footnote{A robot can follow multiple possible trajectories to accomplish a task.} nature of actions. 

In this paper, we propose a new VLA model architecture derived from VLM. Instead of repurposing pretrained VLMs for action prediction, \emph{we use the cognitive information extracted by VLM to guide the  action prediction process of a specialized action module}. To handle the inherent characteristics of action signals -- continuous, multimodal, temporally correlated, and requiring high precision -- we employ advanced diffusion-based transformers (DiT)~\cite{peebles2023scalable} as our action modules, preconditioned on VLM output via the attention mechanism.

The intuition behind our design is \emph{the decoupling  of ``cognition" and ``action" capabilities}. While the large VLMs amass broad visual and semantic knowledge learned from vast amounts of text and images, the cognitive capability and the output language modality have fundamental gaps to dense robot actions. Rather than directly repurposing the VLMs, we advocate the design of componentized VLAs with a dedicated action module.\footnote{As an interesting analogy, our human brain has visual cortex~\cite{visualcortex}, language-relevant cortex~\cite{friederici2011brain}, and motor cortex~\cite{motorcortex} -- the last being dedicated to the control of human body movements.} This action module is specialized for action signal modeling with cognition model output as preconditions. We synergize the cognition and action capabilities via end-to-end training or finetuning. 
Hence, our approach is named \emph{CogACT}.

We systematically study the different backbone architectures for the action module as well as their scalability on model size, and several notable insights have emerged. For example, it is found that sequential modeling with a diffusion transformer significantly outperforms single-step action prediction. More crucially, \emph{we identified a favorable scaling behavior of the action module with diffusion transformers}: adding several hundred million parameters, which is relatively minor compared to a 7B VLM base, results in sizable performance enhancements. This finding suggests the advantages	of a specialized action module and a more efficient approach for VLA model scaling.

In addition to our study on action module design, we also introduce some accompanying algorithms of independent interest. An Adaptive Action Ensemble (AAE) algorithm is proposed to fuse the past action predictions in an adaptive manner, which brings notable performance improvement. We train our VLA models on the Open X-Embodiment dataset~\cite{o2023open_x_embodiment}, and evaluate them on both simulation~\cite{li2024simpler} and  real-robot benchmarks. The comprehensive evaluation and comparisons show that our model performs remarkably well, surpassing existing VLAs by a wide margin.

\vspace{4pt}
\textbf{The contributions of this work} are summarized below:
\begin{itemize}[leftmargin=2em]
	\item We introduce the integration of the action diffusion process into large-scale VLA models.\vspace{2pt}
	\item We propose a componentized VLA model architecture and study the design of large action modules\footnote{Our largest action DiT module is of 300M parameters. Although this may seem modest, especially in comparison with large LLMs/VLMs, it is considered large given the 7D vector space of robot actions we address.} as well as their scaling behaviors.\vspace{2pt}
	\item We propose an adaptive action ensemble algorithm which is simple yet effective for temporal fusion.\vspace{2pt}
	\item Our model achieves significantly better performance than previous VLAs, exhibiting quick adaptation to new robots and tasks and effective generation to unseen objects and backgrounds, as shown in Figure~\ref{fig:teaser}. All our code and models are publicly released.
\end{itemize}

\section{Related Works}

\noindent\textbf{Vision-Language-Action Models.}
The success of Large Language Models (LLMs)~\cite{brown2020language,achiam2023gpt,touvron2023llama,touvron2023llama1} and Vision-Language Models (VLMs)~\cite{liu2024llava,karamcheti2024prismatic, team2023gemini, chen2024internvl, abdin2024phi} has inspired the development of Vision-Language-Action (VLA) models, which extend the capabilities of VLMs by integrating action generation. For instance, RoboFlamingo \cite{li2023vision} extends OpenFlamingo \cite{awadalla2023openflamingo} by incorporating a head network to predict actions and optimizing with MSE loss. RT-2~\cite{brohan2023rt2} tokenizes 7D actions into discrete tokens and uses the VLM PaLI-X~\cite{chen2023pali} to predict them autoregressively like language tokens. OpenVLA adopts a similar approach, tokenizing actions and training the Prismatic VLM~\cite{karamcheti2024prismatic} on the Open-X-Embodiment dataset~\cite{o2023open_x_embodiment}. While these models benefit from VLMs' capabilities and demonstrate promising performance and impressive generalization, they lack the consideration that actions are inherently continuous and temporal, a modality distinct from language.
A group of methods~\cite{wu2023gr1,cheang2024gr2,bharadhwaj2024gen2act} employ large-scale video generative pretraining to enhance visual robot manipulation learning without leveraging pretrained VLMs, and promising results have been demonstrated. 

\vspace{5pt}
\noindent\textbf{Large Action Models.}
There are some recent attempts \emph{concurrent to ours} that explored large action models for generalist robots. 
For example, \cite{hou2024diffusion_transformer_policy} trained a diffusion transformer with 221M parameters and \cite{liu2024rdt} further scaled the action model size to 1B. Both works apply separate vision and language encoders that are pretrained and frozen to process language instructions and images, and they train the action model to integrate these inputs and predict actions with VLA data. Different from ours, these works cannot leverage the generalization and instruction following capability of powerful VLMs pretrained on Internet-scale vision-language aligned data.

\vspace{5pt}
\noindent\textbf{Diffusion-Based Robot Policies.}
Recent studies~\cite{chi2023diffusion_policy,reuss2023goal, pearce2023imitating} have introduced diffusion models as an innovative approach for modeling robotic actions. These diffusion policies have demonstrated strong capabilities to capture the multi-mode nature of robotic action distributions and effectively model the various feasible trajectories that a robot can take to accomplish a given task~\cite{chi2023diffusion_policy}.
Inspired by diffusion policies, Octo~\cite{team2024octo} supplements a transformer-based backbone architecture with compact diffusion heads of 3M parameters to adapt the action output  across different robots. However, the small diffusion head falls short in capturing the precise action distributions and the overall approach does not benefit from strong vision-language models pretrained on web-scale data. In contrast, our work studies large, dedicated action modules (rather than ``heads") with the diffusion transformer architecture. Besides, unlike~\cite{chi2023diffusion_policy, reuss2023goal,pearce2023imitating}, we are interested in large VLAs derived from VLM foundation models with  strong generalization capability.

\section{Method}
\label{sec:method}

\begin{figure*}[!t]
    \centering
    \includegraphics[width=0.99\linewidth]{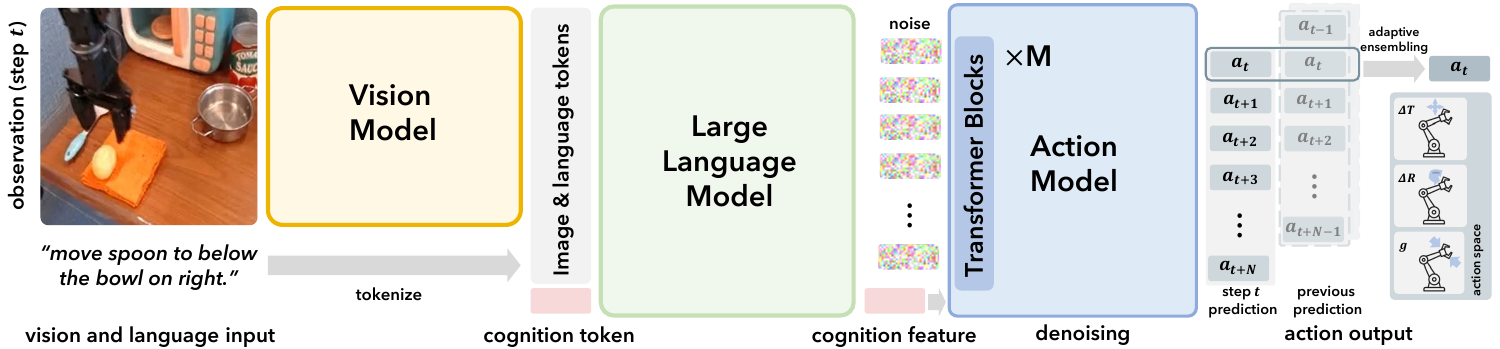}
    \caption{Overview of our architecture. Our model is componentized into three parts: 1) a vision module encoding information from the current image observation into visual tokens; 2) a language module that integrates the visual tokens with the language instructions, and produces a cognition feature determining the desired action to be executed; 3) a diffusion action module, which predicts a sequence of multi-step actions conditioned on the cognition feature. An adaptive ensemble strategy is applied for trajectory ensemble at inference.} 
    \label{fig:method}
\end{figure*}

\paragraph{Problem Formulation.} Our goal is to develop a VLA model that enables different robots to physically execute diverse tasks while receiving visual observations and language instructions. Formally, given the language instruction $\bm{l}$ and visual observation $\bm{o}_t$ at time $t$, a model $\bm{\pi}$ predicts a temporal action sequence $(\bm{a}_t,\bm{a}_{t+1},...,\bm{a}_{t+N})$ for executing the desired task:
\begin{equation}
    \bm{\pi}: (\bm{l},\bm{o}_t) \rightarrow (\bm{a}_t,\bm{a}_{t+1},...,\bm{a}_{t+N}). \label{eq:target}
\end{equation}
While in general, $\bm{a}_t$ can describe various robot actions with different control modes and end-effectors, we consider the action space of a gripper with 7 degrees of freedom (DoF) in this work:
\begin{equation}
    \bm{a}_t = [\Delta x, \Delta y, \Delta z, \Delta \phi, \Delta \theta, \Delta \psi, g], \label{eq:action}
\end{equation}
where $\Delta x, \Delta y, \Delta z$ are the relative translation offsets of the end effector, $\Delta \phi, \Delta \theta, \Delta \psi$ denote the rotation changes, and $g \in \{0,1\}$ indicates the gripper’s open/close state. 

\paravspace
\paragraph{Overall architecture.} To effectively handle complex visual observations and language instructions and collaboratively transform them into precise actions, we componentize the model $\bm{\pi}$ into three parts: a \emph{vision} module, a \emph{language} module, and an \emph{action} module, as shown in Fig.~\ref{fig:method}. 
We describe each part in details below.

\subsection{Vision and Language Modules}
\label{sec:perception}

Our vision and language modules are adapted from an existing VLM from \cite{karamcheti2024prismatic} that has about 7B parameters in total, similar to \cite{kim2024openvla}. We briefly describe them below for completeness.

\paravspace\paragraph{Vision Module.} The vision module processes raw images input into a set of perceptual tokens. It consists of  powerful vision transformers, DINOv2~\cite{oquab2024dinov2} and SigLIP~\cite{zhai2023siglip}, pretrained on Internet-scale image data, to capture rich visual features and a comprehensive semantic understanding of the observations. At each timestep $t$, the image observation $\bm{o}_t$ is fed into the two models, producing two downsampled feature maps $\bm{f}_t^{\mathrm{DINO}}$ and $\bm{f}_t^{\mathrm{Sig}}$, respectively. These feature maps are then concatenated along the channel dimension, passed through a linear projection layer, and serialized into a set of visual perceptual tokens, $\mathcal{V}=\{\bm{v}_1,\bm{v}_2,...,\bm{v}_{N_{\mathcal{V}}}\}$ with a length $N_{\mathcal{V}}$ (we use $256$ by default).

\paravspace\paragraph{Language Module.} The language module is responsible for integrating visual information and language instructions and conducting cognitive reasoning. Here, a LLAMA-2 model~\cite{touvron2023llama} is applied as the backbone.
The language instruction $\bm{l}$ is converted into a set of linguistic tokens, $\mathcal{T}=\{\bm{l}_1,\bm{l}_2,...,\bm{l}_{N_{\mathcal{T}}}\}$, using LLAMA-2's tokenizer. These tokens are then concatenated with the visual tokens $\mathcal{V}$ and an additional learnable cognition token $\bm{c}$, and processed by the model using a causal attention mechanism. The resulting output feature $\bm{f}_t^{\bm c}$, corresponding to the cognition token, encodes integrated information that determines the action to be executed for the current task. This serves as a condition for the subsequent action module to interpret and derive the desired actions. 

\subsection{Diffusion Action Module}
\label{sec:action}

The action module receives the cognition feature as an input condition to generate a series of actions, as defined in Eq.~\eqref{eq:target} and~\eqref{eq:action}. Given that real-world physical actions are continuous and often multi-modal, we predict them using a diffusion modeling process~\cite{nichol2021improved}. To model complex and temporally-correlated actions, we apply a diffusion transformer (DiT)~\cite{peebles2023scalable} as a powerful backbone for the action decoding process.

Specifically, our action module takes the cognition feature $\bm{f}_t^{\bm c}$ along with a series of noisy actions $(\bm{a}_t^i,\bm{a}_{t+1}^i,...,\bm{a}_{t+N}^i)$ as input, where $i$ denotes the current denoising step. It predicts the final actions $(\bm{a}_t,\bm{a}_{t+1},...,\bm{a}_{t+N})$ through multiple denoising steps. The cognition feature and the noisy actions serve as input tokens to the transformer blocks, while the step information $i$ is added to the cognition feature with a sinusoidal positional encoding. 
We enforce the action model to predict not only the current action $\bm{a}_{t}$ but also multiple future actions $(\bm{a}_{t+1},...,\bm{a}_{t+N})$. This approach enhances the overall smoothness of the predicted actions at each time step and increases the final success rates for task execution, as observed similarly in previous studies~\cite{chi2023diffusion_policy,zhao2023act}. 
In practice, the number of predicted future actions is set to a small value ($N=15$ by default), leading to a context length of $N+2=17$ for the action module. This makes the diffusion process highly efficient and does not introduce much computational cost to the overall framework.

\subsection{Training Objective}
Our vision module, language module, and action module are trained/finetuned end-to-end by minimizing the mean-squared error (MSE) between the predicted noises from the action module and the ground truth noises. The loss function is defined as:
\begin{equation}
    \mathcal{L}_{\text{MSE}} = \mathbb{E}_{\bm{\epsilon}\sim\mathcal{N}(0,1),i} ||\boldsymbol{\hat{\epsilon}}^i - \boldsymbol{\epsilon} ||_2, 
\end{equation}
where $\boldsymbol{\hat{\epsilon}}^i$ is the predicted noise for the noisy action sequence $(\bm{a}_t^i,\bm{a}_{t+1}^i,...,\bm{a}_{t+N}^i)$ at the $i$'s denoising step, and $\boldsymbol{\epsilon}$ is the corresponding ground truth.

\subsection{Adaptive Action Ensemble}
\label{sec:inference}

\begin{figure}[!t]
    \centering
    \includegraphics[width=0.99\linewidth]{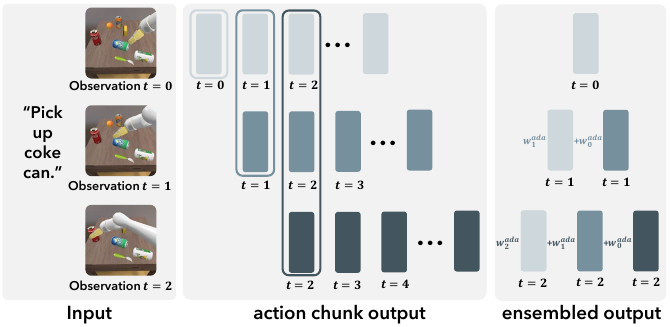}
    \vspace{-2mm}
    \caption{Illustration of our action ensemble strategy with $K=2$ (using the last 2 historical action predictions) as an example.  The action ensemble aggregates historical predictions with the current prediction to jointly determine the final action to be executed.}
    \label{fig:inference}
\end{figure}

During inference, our model predicts actions for multiple time steps. One straightforward strategy is to execute these actions consecutively (Action Chunking~\cite{zhao2023act}) based on the current observation, $\bm{o_t}$. However, this does not fully leverage the visual information available at each time step and may result in jerky motion, as discussed in \cite{zhao2023act}. Alternatively, executing only the action for the current time step (\ie, $\bm{a}_t$) also leads to a less smooth trajectory and diminished performance. 

To alleviate these, \cite{zhao2023act} introduced a temporal ensemble strategy that combines actions predicted for the current time step from both present and past predictions using preset aggregation weights. However, feasible actions for task execution can belong to different modes~\cite{chi2023diffusion_policy}, and simply aggregating them could result in an action that does not align with any mode, which is suboptimal. 

We propose an adaptive ensemble strategy which considers similarities between actions to be aggregated, as shown in Fig.~\ref{fig:inference}. This approach avoid unreasonable aggregation of actions from different modes. Specifically, let $\bm{a}_t|\bm{o}_t$ represent the action prediction for the current time step $t$ given the observation $\bm{o}_t$, and $\{\bm{a}_t|\bm{o}_{t-K},...,\bm{a}_t|\bm{o}_{t-1}\}$ denote the corresponding action predictions based on historical observations $\{\bm{o}_{t-K},...,\bm{o}_{t-1}\}$. We derive the final action $\bm{\hat{a}}_t$ to be executed at time step $t$ as:
\begin{equation}
\label{eq:action_ensemble}
     \hat{\boldsymbol{a}}_t = \sum_{k=0}^{K} w^{\text{ada}}_k \cdot \boldsymbol{a}_{t}|\bm{o}_{t-k}.
\end{equation}
Here, $w^{\text{ada}}_k$ is an adaptive weighting scalar that assigns greater importance to past predictions that are more similar to the current prediction $\bm{a}_t|\bm{o}_t$:
\begin{equation} 
w_k^{\text{ada}} = \text{exp}(\alpha \cdot <\boldsymbol{a}_{t}|\bm{o}_{t},\boldsymbol{a}_{t}|\bm{o}_{t-k}>),
\end{equation}
where $<\cdot,\cdot>$ calculates the cosine similarity between two actions, and $\alpha$ is a hyperparameter set to $0.1$ in practice. 

Our empirical results show that this adaptive action ensemble strategy effectively boosts the success rate of task executions while adding minimal extra cost to inference, since past predictions can be readily cached.

\section{Experiment}

\begin{table*}[!t]
\centering
\small
\caption{Comparison of our approach with existing VLA models on the Google robot across four tasks in two SIMPLER settings. All models are trained on the Open X-Embodiment dataset, except for RT-1 which is trained exclusively on the Google robot subset.}
\vspace{-4pt}
\begin{tabular}{l|l|cccc|c}
\toprule
\multirow{2}{*}{Google Robot} &
  \multirow{2}{*}{Method} &
  Pick &
  Move &
  Open/Close & Open Top Drawer & \multirow{2}{*}{\emph{Average}} \\ 
  & & Coke Can & Near & Drawer & and Place Apple & \\
  \midrule
\multirow{6}{*}{\begin{tabular}[l]{@{}l@{}}SIMPLER \\ (Visual Matching) \end{tabular}}
             &  \textcolor{gray}{RT-1~\cite{brohan2022rt1}}  &  \textcolor{gray}{85.7}     &  \textcolor{gray}{44.2}     &  \textcolor{gray}{73.0}     &   \textcolor{gray}{~~6.5}    &  \textcolor{gray}{52.4}   \\
             & RT-1-X~\cite{o2023open_x_embodiment}      & 56.7     & 31.7     & 59.7     & 21.3    & 42.4  \\ 
             & RT-2-X~\cite{o2023open_x_embodiment}   & 78.7   &  77.9  & 25.0  &  ~~3.7 & 46.3  \\ 
             & Octo-Base~\cite{team2024octo}      & 17.0     & ~~4.2    & 22.7     &  ~~0.0    & 11.0    \\ 
             & OpenVLA~\cite{kim2024openvla}      & 18.0     & 56.3     & 63.0     & ~~0.0     & 34.3   \\
             & Ours & \textbf{91.3}     & \textbf{85.0}     & \textbf{71.8}    &  \textbf{50.9}   & \textbf{74.8}  \\
             \midrule
\multirow{6}{*}{\begin{tabular}[l]{@{}l@{}}SIMPLER \\ (Variant Aggregation) \end{tabular}}
             & \textcolor{gray}{RT-1~\cite{brohan2022rt1}}  & \textcolor{gray}{89.8}     & \textcolor{gray}{50.0}    & \textcolor{gray}{32.3}    &  \textcolor{gray}{~~2.6}    & \textcolor{gray}{43.7}  \\
             & RT-1-X~\cite{o2023open_x_embodiment}     & 49.0    & 32.3    & 29.4     &  10.1    & 30.2    \\ 
             & RT-2-X~\cite{o2023open_x_embodiment}  & 82.3  &  79.2 & \textbf{35.3} & 20.6 & 54.4 \\ 
             & Octo-Base~\cite{team2024octo}     & ~~0.6     & ~~3.1    & ~~1.1     &  ~~0.0    & ~~1.2    \\ 
             & OpenVLA~\cite{kim2024openvla}  & 60.8     & 67.7     & 28.8     &  ~~0.0   &  39.3    \\
             & Ours  & \textbf{89.6}     & \textbf{80.8}     & 28.3     & \textbf{46.6}     & \textbf{61.3}      \\    
\bottomrule
\end{tabular}

\label{tab:google_robot}
\end{table*}

\begin{table*}[!t]
\centering
\small
\caption{Evaluation results on the WidowX robot in the SIMPLER \textit{Visual Matching} setting. 
For these tests, we repeat each task 5 times to improve the statistical significance (see \emph{suppl. material} for details), and thus the results of Octo may slightly differ from those in \cite{li2024simpler}.
}
\vspace{-4pt}
\begin{tabular}{l|l|cccc|c}
\toprule
\multirow{2}{*}{ WidowX Robot} &
  \multirow{2}{*}{Method} &
  Put Spoon  &
  Put Carrot  &
   Stack Green Block  & Put Eggplant  & \multirow{2}{*}{\emph{Average}} \\ 
   & &on Towel & on Plate  & on Yellow Block & in Yellow Basket  &   \\
   \midrule
\multirow{5}{*}{\begin{tabular}[l]{@{}l@{}}SIMPLER\\ (Visual Matching) \end{tabular}} 
             & RT-1-X~\cite{o2023open_x_embodiment}     & ~~0.0     & ~~4.2     & ~~0.0     & ~~0.0    & ~~1.1  \\ 
             & Octo-Base~\cite{team2024octo}  & 15.8   &  12.5  & ~~0.0  &  41.7 & 17.5 \\ 
             & Octo-Small~\cite{team2024octo}     & 41.7     & ~~8.2    & ~~0.0     &  56.7    & 26.7    \\ 
             & OpenVLA~\cite{kim2024openvla}     & ~~4.2     & ~~0.0     & ~~0.0     & 12.5     & ~~4.2   \\
             & Ours & \textbf{71.7}     & \textbf{50.8}     & \textbf{15.0}      &  \textbf{67.5}    & \textbf{51.3}  \\
\bottomrule
\end{tabular}

\label{tab:widowx}
\end{table*}

\noindent\textbf{Training Dataset.} We use the Open X-Embodiment (OXE)~\cite{o2023open_x_embodiment} dataset as our primary training dataset. It includes over 1 million real-world robotic trajectories collected from 60 datasets, covering 22 different robot embodiments. 
We use the similar subset of OXE as in Octo~\cite{team2024octo} and OpenVLA~\cite{kim2024openvla} for training, which comprises 22.5 million frames. 
For details regarding data distributions, please refer to~\cite{team2024octo,kim2024openvla}.

\vspace{5pt}
\noindent\textbf{Implementation Details.} The model is trained with a batch size of 256 and 8 diffusion steps per sample, initialized with pre-trained vision and language module weights from \cite{kim2024openvla}. The vision module (\ie, DINOv2 and SigLIP), language module (\ie, LLAMA-2), and action module are all trained end-to-end, following a constant learning rate of $2e-5$ for over 135K iterations. Training is conducted on 16 NVIDIA A100 GPUs with approximately 5 days using PyTorch’s Fully Sharded Data Parallel (FSDP) framework. By default, we use DiT-Base as our action model. The ensemble window $K$ is set to be inversely proportional to the moving distance per frame, which can be inferred from robots' motion speed and observation frequency. In practice, we use the standard deviation of actions from training set to determine $K$, \ie, $2$ for RT-1 dataset with Google Robot, $7$ for BridgeDataV2 with WidowX Robot. 
\subsection{Simulated Evaluation}
\label{subsec:sim_eval}
\noindent\textbf{Evaluation Environment.} After training, we evaluate our model within the SIMPLER~\cite{li2024simpler} evaluation environment. This simulation platform is designed to bridge the real-to-sim control and visual gap by faithfully replicating real-world conditions for robots like the Google robot and the WidowX robot. Extensive testing of various VLA models has shown a strong correlation between performance in SIMPLER and real-world outcomes~\cite{li2024simpler}.

\vspace{5pt}
\noindent\textbf{Evaluation Settings.} SIMPLER offers two evaluation settings: \textit{Visual Matching}, which closely replicates real-world tasks by minimizing discrepancies between the simulated and real environments, and \textit{Variant Aggregations}, which introduces variations to \textit{Visual Matching} by modifying elements such as background, lighting, distractors, and table texture. For the Google robot, SIMPLER provides both the two evaluation settings, each featuring the same four tasks: 1) Pick coke can; 2) Move near; 3) Open/close drawer; and 4) Open top drawer and place apple. For the WidowX robot, SIMPLER provides only the \textit{Visual Matching} setting, with four tasks: 1) Put spoon on towel; 2) Put carrot on plate; 3) Stack green block on yellow block; and 4) Put eggplant in yellow basket. Success rate is used as the evaluation metric.

\vspace{5pt}
\noindent\textbf{Experiments on Google Robot.}  
Table~\ref{tab:google_robot} shows the success rates of our model and compares our approach with existing VLA models on the Google robot across four tasks in both SIMPLER settings. Our model achieves the highest average success rate in both settings, with 74.8\% in \textit{Visual Matching} and 61.3\% in \textit{Variant Aggregation}. Remarkably, our model even outperforms RT-1~\cite{brohan2022rt1}, which is trained on a Google robot-specific dataset, by an average success rate margin of 22.4\% in \textit{Visual Matching} and 17.6\% in \textit{Variant Aggregation}. Additionally, the average success rate of our model  significantly surpasses that of RT-2-X~\cite{o2023open_x_embodiment}, despite our model being much smaller, with 7.6B parameters compared to RT-2-X that has 55B parameters.

\vspace{5pt}
\noindent\textbf{Experiments on WidowX Robot.} 
Table~\ref{tab:widowx} presents the evaluation results of our model compared with the other methods on the WidowX robot using the SIMPLER environment in the \textit{Visual Matching} setting. Our model also attains the highest average success rate of 51.3\%, outperforming the other models by a significant margin.

\subsection{Real-World Evaluation with Realman Robot}
\label{sec:realman}
\noindent\textbf{Robot.} We perform real-world experiments using the Realman Arm\footnote{https://www.realman-robotics.com/rm75-b.html}, which has 7 DoFs and is equipped with a 1-DoF gripper. Inverse kinematics is applied to determine joint angles, enabling the end effector to align with the model’s predicted pose.
\vspace{5pt}
\noindent\textbf{Task Definitions.} We design three tasks for real-world experiments, named ``Pick'', ``Stack'', and ``Place'':
\begin{itemize}
    \item Pick up the \texttt{Object} and place it onto the \texttt{Color} plate, where $\texttt{Object} \in \{\text{Banana}, \text{Lemon}, \text{Avocado}\}$, and $\texttt{Color} \in \{\text{White}, \text{Blue}, \text{Yellow}\}$.
    \item Stack the \texttt{Color} \texttt{Object} into the \texttt{Color} \texttt{Object}, where $\texttt{Object} \in \{\text{Cup}, \text{Bowl}\}$ and $\texttt{Color} \in \{\text{Pink}, \text{White}, \text{Blue}, \text{Yellow}\}$.
    \item Place the \texttt{Color} block onto the \texttt{Color} block, where $\texttt{Color} \in \{\text{Red}, \text{Orange}, \text{Blue}, \text{Green}, \text{Yellow}\}$.
\end{itemize}

\vspace{5pt}
\noindent\textbf{Fine-Tuning Data.} We manually collect 48, 67 and 79 demonstrations for the ``Pick'', ``Stack'', and ``Place'' tasks, respectively. Additionally, we capture 197 demonstrations for various other tasks like ``Pick up the orange can and drop it into the basket''. In total, we gather 391 demonstrations for fine-tuning. Notably, demonstrations for tasks identical to the evaluation settings are very few. For example, there are only two demonstrations for the specific task ``put the banana on the yellow plate.''
Details of the fine-tuning process are provided in the \emph{suppl. material}.

\begin{figure*}[!t]
    \centering
    \includegraphics[width=0.98\linewidth]{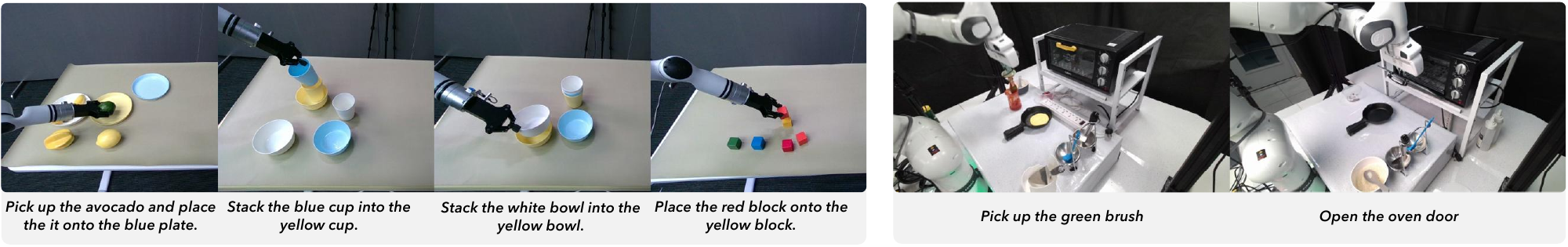}
    \vspace{-4pt}
    \caption{Real-world evaluation environments of Realman robot (left) and Franka robot (right).
    } 
    \label{fig:realman}
\end{figure*}

\begin{table*}[!t]
\centering
\small
\caption{Real-world evaluation with the Realman Robot across three tasks. All models are pre-trained on OXE and then fine-tuned on our collected data.}
\vspace{-4pt}
\begin{tabular}{l|cccc|ccc|ccc|c}
\toprule
  \multirow{2}{*}{Method} & \multicolumn{4}{c|}{Pick}  & \multicolumn{3}{c|}{Stack}  & \multicolumn{3}{c|}{Place} & Task (All) \\ \cmidrule(lr){2-5} \cmidrule(lr){6-8} \cmidrule(lr){9-11} \cmidrule(lr){12-12}
 & Banana & Lemon & Avocado &  Avg. & Cup & Bowl & Avg. & Pick & Stack & Avg. &  Avg. \\  
   \midrule
             Octo-Base~\cite{team2024octo}     &  25.0   & ~~0.0    & ~~0.0   &  ~~8.3  & ~~0.0 & ~~0.0 & ~~0.0& 12.5 & ~~0.0 & ~~6.3 & ~~4.9 \\
 
             OpenVLA~\cite{kim2024openvla}   &  12.5  &12.5  &  ~~0.0 & ~~8.3 &25.0 &  ~~6.3  & 15.6      &25.0 & ~~4.2 & 12.5& 12.1 \\
             Ours       & \textbf{75.0}  & \textbf{50.0}   & \textbf{87.5}    &\textbf{70.8} &  \textbf{95.8} & \textbf{68.8} & \textbf{82.3} & \textbf{87.5} & \textbf{33.3} & \textbf{60.4}& \textbf{71.2} \\
\bottomrule
\end{tabular}

\label{tab:real-world}
\end{table*}

\begin{table*}[!t]
\centering
\small
\caption{Real-world generalization evaluation with the Realman Robot on \emph{unseen} tables with additional \emph{unseen} distractors.}
\vspace{-4pt}
\begin{tabular}{l|cccc|ccc|ccc|c}
\toprule
  \multirow{2}{*}{Method} & \multicolumn{4}{c|}{Pick}  & \multicolumn{3}{c|}{Stack}  & \multicolumn{3}{c|}{Place} & Task (All) \\ \cmidrule(lr){2-5} \cmidrule(lr){6-8} \cmidrule(lr){9-11} \cmidrule(lr){12-12}
 & Banana & Lemon & Avocado &  Avg. & Cup & Bowl & Avg. & Pick & Stack & Avg. &  Avg. \\  
   \midrule
             OpenVLA~\cite{kim2024openvla}   &  12.5  & ~~0.0 & 12.5  & ~~8.3 &12.5 & ~~0.0  & ~~6.3     &29.2 & ~~0.0 & 14.6& ~~9.7 \\
             Ours       & \textbf{75.0}  &\textbf{62.5}   & \textbf{87.5}    &\textbf{75.0} &  \textbf{83.3} & \textbf{45.8} & \textbf{64.6} &\textbf{54.2} & \textbf{16.7} & \textbf{35.5} & \textbf{58.4} \\
\bottomrule
\end{tabular}

\label{tab:background generation and distractors}
\end{table*}

\vspace{5pt}
\noindent\textbf{Evaluation.} 
Each task is evaluated as follows:
\begin{itemize}
    \item ``Pick'': 24 evaluations in total, with 8 trials per object (Banana, Lemon and Avocado). We report the success rate for each object, as well as the average success rate across the three objects.
    \item ``Stack'': 48 evaluations in total, with 24 trials per object (Cup and Bowl). We report the success rate for each object, along with the average success rate across both objects.
    \item ``Place'': 24 evaluations in total. This task is divided into two steps: 1) picking up the block, and 2) placing the block onto another block. We report the success rate for each step, and the average success rate across both steps. 
\end{itemize}

\begin{table}[!t]
\centering
\small
\caption{Real-world generalization evaluation with the Realman Robot on unseen colors, shapes and categories.}
\vspace{-4pt}
\begin{tabular}{lcccc}
\toprule
  \multirow{2}{*}{Method} &
   Unseen &
   Unseen  &
   Unseen & \multirow{2}{*}{Avg.}\\ 
   
 & Colors & Shapes & Categories & \\ \midrule
             OpenVLA~\cite{kim2024openvla}     & ~~0.0     & ~~6.3    & 12.5     & ~~6.3      \\ 
             Ours  & \textbf{87.5}   & \textbf{81.3} & \textbf{25.0}  &  \textbf{64.6}  \\ 
\bottomrule
\end{tabular}

\label{tab:unseen colors}
\end{table}

In each evaluation, we randomly choose the \texttt{Color}, and place both the target and at least three distractor objects in random positions on the table. We vary the table height within the robot's accessible range, and randomly set the robot’s position around the table.

\vspace{5pt}
\noindent\textbf{Results.} Table~\ref{tab:real-world} presents a comparison of our model with Octo-Base~\cite{team2024octo} and OpenVLA~\cite{kim2024openvla}. For a fair evaluation, all models are pre-trained on the OXE dataset and subsequently fine-tuned using our collected demonstrations. Our model achieves a notable improvement, outperforming OpenVLA by a margin of 59.1\% in success rate.

\vspace{5pt}
\noindent\textbf{Generalization Evaluation.} We assess the generalization capability of the fine-tuned models on environments, distractors, tables, and objects that are not present in the fine-tuning dataset.
We exclude Octo-Base~\cite{team2024octo} from this evaluation since its success rate in the seen task is already close to zero.
\begin{itemize}
    \item \textit{Unseen Tables with Unseen Distractors.} The table color is altered to closely resemble the colors of the background and floor, increasing visual complexity. Additionally, three new unseen objects are introduced as distractors. The results for each task is presented in Table~\ref{tab:background generation and distractors}.
    \item \textit{Unseen Colors, Shapes, and Categories.} We assess performance on unseen colors in the ``Pick'' task, using seen objects (Banana, Avocado) with new colors (Green, Pink) of plates. Each object-color combination is evaluated twice, for a total of 8 trials. 
    
    For evaluating on unseen shapes, we define four new tasks: (1-2) ``Pick up the \texttt{Shape} block and put it into the basket'', where $\texttt{Shape} \in \{\text{Triangular}, \text{Arched}\}$; (3) ``Pick up a small can and drop it into the basket''; (4) ``Stack one cylindrical block on top of another cylindrical block''. Each task is evaluated 4 times, resulting in a total of 16 trials. 
    
    For evaluating unseen categories, we introduce a new task: ``Pick up the \texttt{Object} and place it into the basket'', where $\texttt{Object} \in \{\text{Eggplant}, \text{Hammer}\}$. This task is evaluated 8 times in total, with 4 trials per object.

    Table~\ref{tab:unseen colors} presents the results, demonstrating that our approach exhibits strong generalization capabilities, particularly in handling unseen colors and shapes.
\end{itemize}

 \begin{table}[!t]
\centering
\small
\caption{Real-world evaluation with the Franka Robot across four tasks. All models are pre-trained on OXE and then fine-tuned on our collected data.
}
\vspace{-4pt}
\begin{tabular}{lcccccc}
\toprule
  \multirow{2}{*}{Method} &
   Close&
   Open  &
   Pick & Pick & \multirow{2}{*}{Avg.}\\ 
   
 & Oven & Oven & Bowl & Brush \\ \midrule
            Octo-Base~\cite{team2024octo}     & ~~0.0    & ~~0.0    & 27.3     &  ~~0.0 &  ~~5.8  \\ 
             OpenVLA~\cite{kim2024openvla}     & 18.2     & ~~0.0    & ~~9.1     & ~~0.0  & ~~6.8    \\ 
             Ours  & \textbf{63.6}   & \textbf{72.7} & \textbf{72.7}  &  \textbf{36.4}  & \textbf{61.4} \\ 
\bottomrule
\end{tabular}
\label{tab:Franka Emilka Exp}
\end{table}

\subsection{Real-World Evaluation with Franka Robot}
\label{sec:Franka}
\noindent\textbf{Robot.} To further demonstrate the generalization capability of our approach, we employ an additional robot, the Franka Arm\footnote{https://franka.de/}, which features 7 DoFs and a 1-DoF gripper, for real-world evaluation.

\vspace{5pt}
\noindent\textbf{Task Definitions.} We define four tasks for evaluation: 1) Close the oven door; 2) Open the oven door; 3) Pick up the green brush; 4) Pick up a bowl containing food.

\vspace{5pt}
\noindent\textbf{Fine-Tuning Data.} For each task, we collect 100 demonstration samples, resulting in a total of 400 samples used to fine-tune all the models.

\vspace{5pt}
\noindent\textbf{Results.} Each task is tested over 11 trials. Table~\ref{tab:Franka Emilka Exp} presents a comparison between our model, Octo-Base, and OpenVLA, demonstrating that our approach achieves significantly higher performance. Real-world results with the Realman Robot (Table~\ref{tab:real-world}) and the Franka Robot (Table~\ref{tab:Franka Emilka Exp})  consistently validate the generalization capability of our approach across different robotic platforms.

\subsection{Ablation Study}
\label{sec:ablations}
We employ SIMPLER evaluation on both the Google robot and the WidowX robot for all ablation studies. We use the following abbreviations: GR for Google Robot, WR for WidowX Robot, VM for the SIMPLER Visual Matching setting, and VA for the SIMPLER Visual Aggregation setting.

\vspace{5pt}
\noindent\textbf{Action Model Architectures.} 
We evalute various action model architectures on GR and WR. The architectures examined include Multi-Layer Perceptrons (MLPs) with depths of 3 and 7 layers, respectively, as well as a series of diffusion transformers of varying scales. The hidden state dimensions are set to 256 and 1024 for the two MLPs. 

Table~\ref{tab:action_structure} shows that both MLP and transformer structures show improved success rate with increased model size. With the same number of parameters, transformers outperform MLPs, likely due to the attention mechanism's superior sequence modeling capabilities. Notably, DiT-Large achieves the highest average success rate of 64.8\%. As shown in Figure~\ref{fig:teaser}, the average success rate of transformers is approximately linearly related to the logarithm of the model size. This indicates a promising scaling behavior of the action module with diffusion transformers.

\begin{table}[!t]
\centering
\small
\caption{Performance comparison of different action model network structures. }
\vspace{-4pt}
\begin{tabular}{l|c|ccc|c}
\toprule
  \multirow{2}{*}{Action Model}  &\multirow{2}{*}{\!Params\!} &
   \multicolumn{2}{c}{GR}   &
   WR   &  \multirow{2}{*}{Average}\\
   & &(VM)  & (VA)& (VM)&  \\ 
      \midrule
              MLP (3-Layer)   & 3M & 52.2     &  52.4     & 47.1    & 50.6    \\
             MLP (7-Layer)  & 89M  & 61.4     & 48.0    & 48.1     &  52.5     \\ 
             DiT-Small & 13M &  73.3  & 51.3   & 51.0  &  58.5  \\ 
             DiT-Base   & 89M & \underline{74.8}     & \textbf{61.3}     & \underline{51.3}     & \underline{62.5}     \\ 
              DiT-Large   & 308M & \textbf{76.7}     & \underline{59.3}     & \textbf{58.3}     & \textbf{64.8}     \\ 
\bottomrule
\end{tabular}

\label{tab:action_structure}
\end{table}

\begin{table}[!t]
\centering
\small
\caption{Impact of multi-step action prediction on model performance. A future step setting of 0 indicates no multi-step action prediction during training.}
\vspace{-4pt}
\begin{tabular}{l|ccc|c}
\toprule
   \multirow{2}{*}{Future Steps}&
   \multicolumn{2}{c}{GR}   &
   WR   &  \multirow{2}{*}{Average}\\
      &(VM) & (VA)& (VM)&  \\ 
      \midrule
             0     & 73.4     & 49.0     & ~~6.3     & 42.8     \\
            3  & 70.4   &  58.9 & 37.1  &  55.5  \\ 
            15     & \textbf{74.8}     & \textbf{61.3}     & 51.3     & \textbf{62.5}     \\
            31 & 54.3   &  47.6 &  \textbf{51.7} & 51.2   \\ 
\bottomrule
\end{tabular}

\label{tab:action_step}
\end{table}

\vspace{5pt}
\noindent\textbf{Multi-Step Action Prediction.} During training, our model predicts the action for the current step along with $N$ future steps. In Table~\ref{tab:action_step}, we examine the effect of varying $N$ (0, 3, 15 and 31 future steps) on performance. 
Our results indicate that predicting 15 future steps leads to the highest performance.

\vspace{5pt}
\noindent\textbf{Adaptive Action Ensemble.} 
In Section~\ref{sec:inference}, we present an action ensemble strategy, termed Adaptive Ensemble, as formulated in Eq.~\eqref{eq:action_ensemble}. We evaluate this approach against the two ensemble strategies introduced in ACT~\cite{zhao2023act}: Action Chunking and Temporal Ensemble.
Given a sequence of action predictions of current observation, Action Chunking executes the first two predictions directly. We use the same ensemble windows for Temporal Ensemble and our Adaptive Ensemble.
Table~\ref{tab:action_ensemble} presents the success rates of these strategies. Our proposed Adaptive Ensemble outperforms others, and we attribute this to its integration of similarity weighting between current and historical predictions.

\begin{table}[!t]
\centering
\small
\caption{Comparison of our proposed action ensemble strategy, Adaptive Ensemble, against two strategies introduced in \cite{zhao2023act}.
}
\vspace{-4pt}
\begin{tabular}{l|ccc|c}
\toprule
   \multirow{2}{*}{Strategy}&
   \multicolumn{2}{c}{GR}   &
   WR   &  \multirow{2}{*}{Average}\\
      &(VM) & (VA)& (VM)&  \\ 
      \midrule
             Action Chunking     & 67.4     & 52.5     & 32.1     & 50.7     \\ 
             Temporal Ensemble  & \textbf{75.0}   &  59.9 & 41.9 & 58.9   \\ 
              Adaptive Ensemble    & 74.8     & \textbf{61.3}     & \textbf{51.3}      & \textbf{62.5}   \\
\bottomrule
\end{tabular}
\vspace{-8pt}
\label{tab:action_ensemble}
\end{table}
\section{Conclusion}
We have presented a large VLA model designed with a specialized focus on action modeling. Unlike previous VLAs that employ simple adaptations of vision-language models for action prediction, CogACT separates cognitive and action capabilities, using advanced diffusion transformers as a dedicated action module. This approach effectively addresses the continuous, multimodal, and temporally correlated nature of robot actions, leading to substantial improvements in performance and generalization. Our findings highlight the advantages of a componentized VLA model, where a large VLM serves as the cognitive foundation while the DiT action module handles precise, sequential action prediction. Favorable scaling behaviors of the action module have been observed,  where modest parameter increases yield significant performance gains. The extensive experiments show that our model not only significantly surpasses existing VLAs in task performance and but also exhibits remarkable generation capability to unseen objects and backgrounds.

{
    \small
    \bibliographystyle{ieeenat_fullname}
    \bibliography{main}
}
\newpage
\maketitlesupplementary

  
  



\renewcommand{\thesection}{\Alph{section}}
\renewcommand{\thefigure}{\Roman{figure}}
\renewcommand{\thetable}{\Roman{table}}
\renewcommand{\theequation}{\Roman{equation}}
\setcounter{section}{0}
\setcounter{table}{0}
\setcounter{figure}{0}
\setcounter{equation}{0}

\appendix
\section{Training Data Details}

\subsection{Pretraining Data}
\paragraph{Vision-Language Data.} Our VLA model is built upon a pretrained VLM, Prismatic~\cite{karamcheti2024prismatic}. Here, we briefly describe the training data of \cite{karamcheti2024prismatic} for self-containedness purposes and refer the readers to \cite{karamcheti2024prismatic,oquab2024dinov2,zhai2023siglip,touvron2023llama} for more details. The Prismatic model uses DINOv2~\cite{oquab2024dinov2} and SigLIP~\cite{zhai2023siglip} as the vision modules, which were trained with 1.2 billion images and 40 billion image-text pairs, respectively. A LLaMa-2~\cite{touvron2023llama} is employed as the LLM backbone, which was trained on 2 trillion language tokens. The vision and LLM modules were further finetuned with 1.2 million multimodal instruct tuning examples by \cite{karamcheti2024prismatic}.

\paravspace\paragraph{Vision-Language-Action Data.} We pretrain our model on 25 VLA datasets from Open X-Embodiment~\cite{o2023open_x_embodiment}. 
As in Octo~\cite{team2024octo} and OpenVLA~\cite{kim2024openvla}, we restrict our training on datasets with single-arm end-effector control and at least one third-person camera perspective. 
Our data mixture strategy primarily follows \cite{team2024octo,kim2024openvla}, except that we do not use the Language Table~\cite{lynch2023interactive} and Droid~\cite{khazatsky2024droid} datasets in our entire training process due to their significant distribution disparities with other data.
The detailed data mixture is listed in Table~\ref{tab:data_mix}. 
In total, we use 0.4 million robot trajectories containing 22.5 million frames as our training data.

\begin{table}[th]
    \centering
        \caption{Our training data mixture using datasets from the Open X-Embodiment dataset~\cite{o2023open_x_embodiment}.}
       \begin{tabular}{lr}
        \toprule
        Dataset & Ratio \\
        \midrule
        Fractal~\citep{brohan2022rt} & 27.1\% \\
        Kuka~\citep{kalashnikov2018qt} & 14.7\% \\
        Bridge~\citep{ebert2021bridge, walke2023bridgedata} & 15.3\% \\
        Taco Play~\citep{rosete2022tacorl, mees23hulc2} & 3.4\% \\
        Jaco Play~\citep{dass2023jacoplay} & 0.6\% \\
        Berkeley Cable Routing~\citep{luo2023multistage} & 0.3\% \\
        Roboturk~\citep{mandlekar2019scaling} & 2.7\% \\
        Viola~\citep{zhu2022viola} & 1.1\% \\
        Berkeley Autolab UR5~\citep{BerkeleyUR5Website} & 1.4\% \\
        Toto~\citep{zhou2023train} & 2.3\% \\
        Stanford Hydra Dataset~\citep{belkhale2023hydra}  & 5.1\% \\
        Austin Buds Dataset~\citep{zhu2022bottom}  & 0.2\% \\
        NYU Franka Play Dataset~\citep{cui2022play}  & 1.0\% \\
        Furniture Bench Dataset~\citep{heo2023furniturebench}  & 2.8\% \\
        UCSD Kitchen Dataset~\citep{ucsd_kitchens}  & \textless 0.1\% \\
        Austin Sailor Dataset~\citep{nasiriany2022sailor}  & 2.5\% \\
        Austin Sirius Dataset~\citep{liu2022robot}  & 2.0\% \\
        DLR EDAN Shared Control~\citep{quere_shared_2020}  & \textless 0.1\% \\
        IAMLab CMU Pickup Insert~\citep{saxena2023multiresolution}  & 1.0\% \\
        UTAustin Mutex~\citep{shah2023mutex} & 2.6\% \\
        Berkeley Fanuc Manipulation~\citep{fanuc_manipulation2023} & 0.9\% \\
        CMU Stretch~\citep{mendonca2023structured} & 0.2\% \\
        BC-Z~\citep{jang2022bc} & 8.6\% \\
        FMB Dataset~\citep{luo2024fmb}  & 2.4\% \\
        DobbE~\citep{shafiullah2023dobbe}  & 1.6\% \\

                \bottomrule
        \end{tabular}%
        \label{tab:data_mix}
\end{table}

\subsection{Finetuning Data for Real Robot Experiments}
\begin{figure}[!t]
    \centering
    \includegraphics[width=0.98\linewidth]{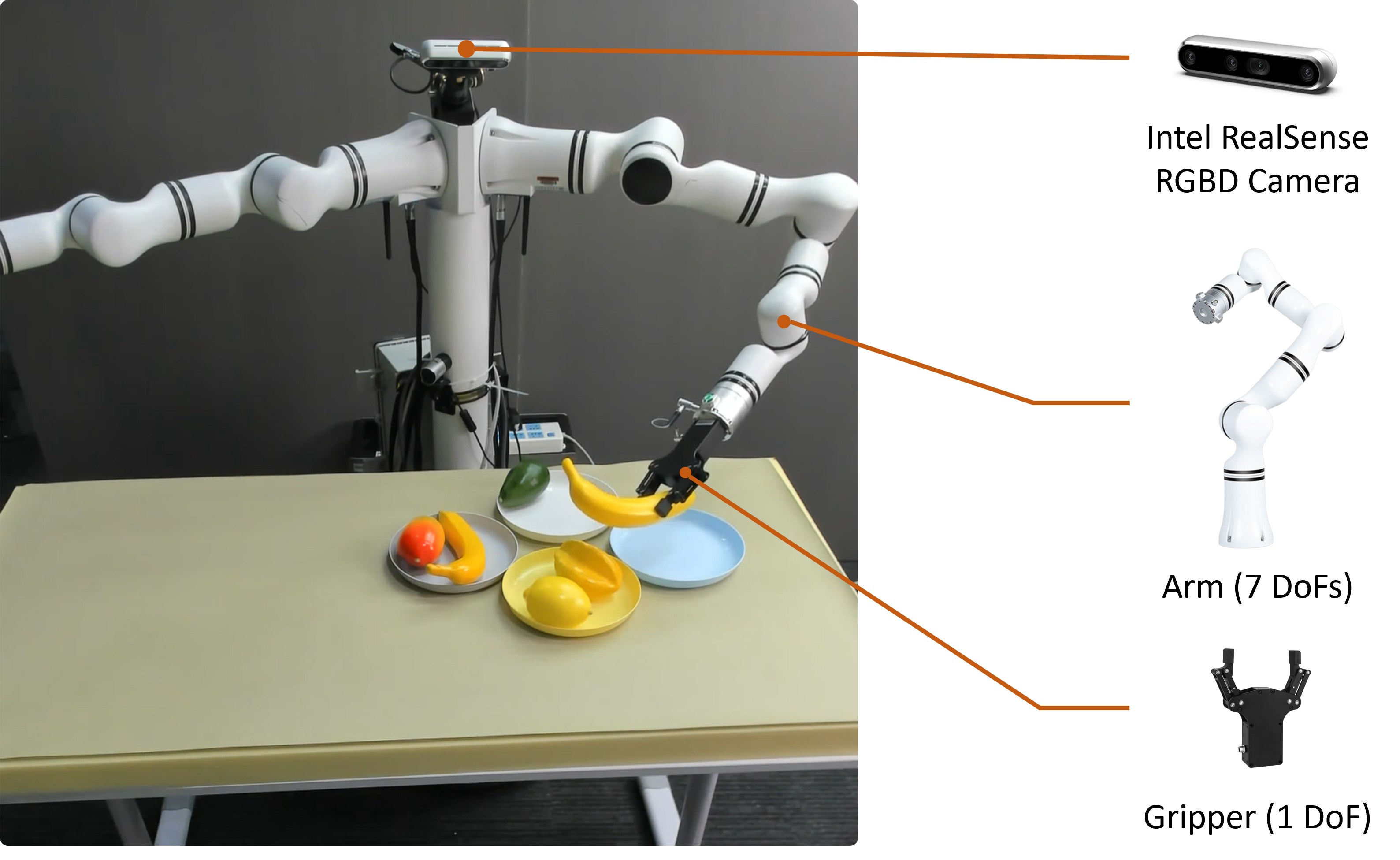}
    \vspace{-8pt}
    \caption{Realman robot setup (right arm not used).
    } 
    \label{fig:realman_robot}
\end{figure}

\paragraph{Realman Robot Setup.} 
Our hardware setup for the Realman robot experiments is presented in Figure~\ref{fig:realman_robot}. We built a robot with two Realman arms on its shoulders, each having 7 degrees of freedom (DoF). We connect the left arm with a 1-DoF gripper and only use this arm in all experiments. We use Intel RealSense camera to capture the RGB image. We perform hand-eye calibration each time we move the camera. We convert all actions in the training data to the camera coordinate system, and the actions generated by the model to the robot coordinate system. To evaluate the generalization ability of our model across different viewpoints, we randomly move the camera before our experiments. 
Our robot is equipped with a mobile base, which was repositioned randomly before each experiment in order to assess the model's generalization ability.

\begin{figure}[!t]
    \centering
    \includegraphics[width=0.98\linewidth]{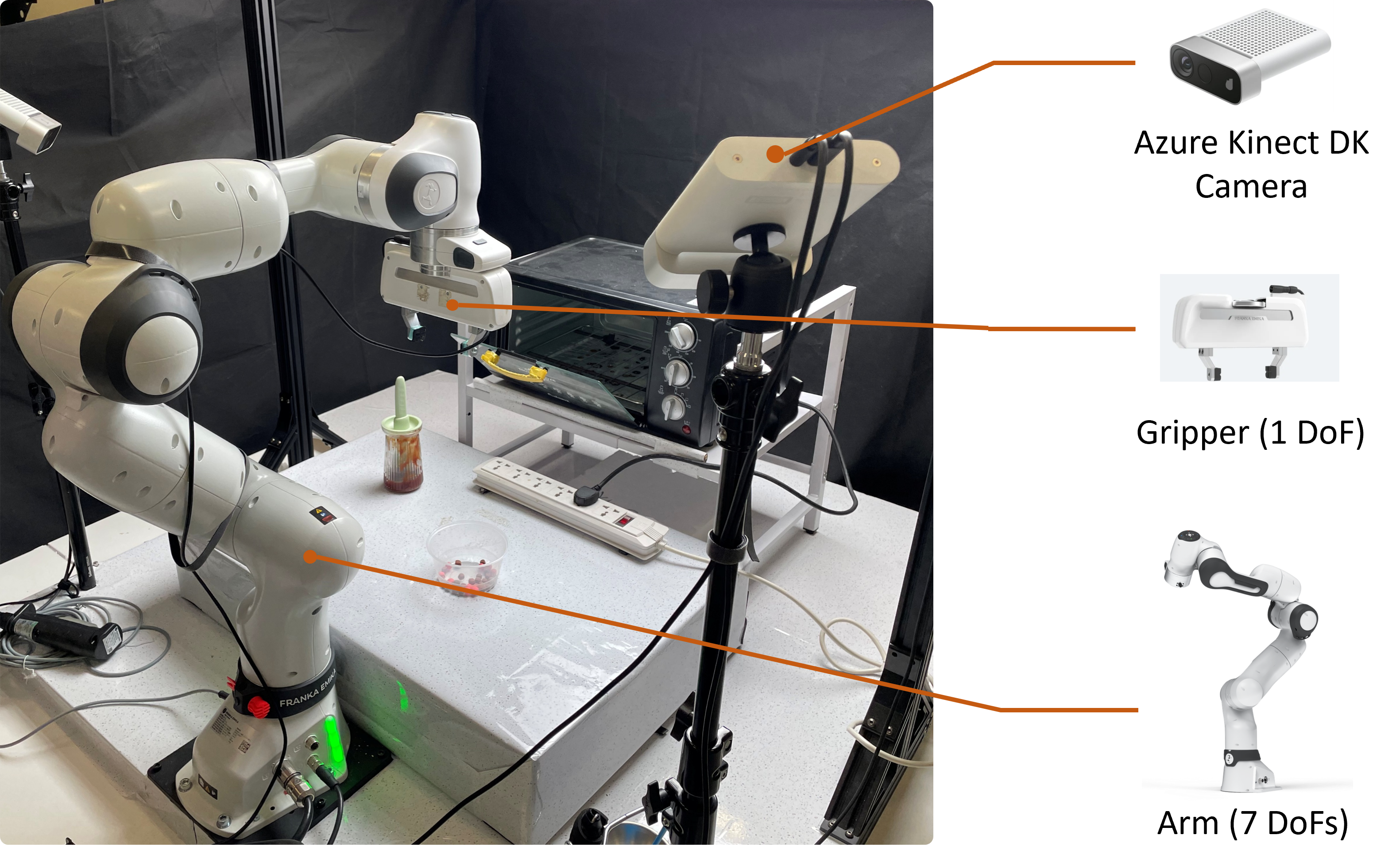}
    \vspace{-8pt}
    \caption{Franka robot setup.
    }
    \label{fig:franka_robot}
\end{figure}

\paravspace\paragraph{Franka Robot Setup.} 
The Franka Robot setup is shown in Figure~\ref{fig:franka_robot}. For this embodiment, a robot arm which has 7 DoF is rigidly attached onto a table. We use a Kinect DK camera on the right to capture RGB images. The actions throughout the entire pipeline are in the robot coordinate system. 

\paravspace\paragraph{Data Collection.} 
We use the touch controller from a Meta Quest 2 device to teleoperate the robot and collect demonstration data for both robots. The translation and rotation of the touch controller are mapped to the gripper's 3D motion, and a button controls the opening and closing of the gripper. We record video frames at 30 Hz for the Realman robot and 5$\sim$6 Hz for the Franka robot. We manually trim the video segments from the recorded full videos and append the language instructions describing the tasks.

\paravspace\paragraph{Data Preprocessing.} 
We follow the format of the Open X-Embodiment dataset~\cite{o2023open_x_embodiment} to process both the two fine-tuning datasets of the Realman robot and Franka robot. We crop and resize the image to 224\texttimes 224, and transform the actions to relative translation offset and rotation changes. We calculate the rotation changes in rotation matrix and transform them to Euler angle for training. During training, the Realman robot data is randomly subsampled to 5 Hz with a time interval of 0.2 seconds between frames. The Franka robot data is used as is.

\subsection{Data Augmentation and Normalization}
We utilize data augmentations for images like random crop, random brightness, random contrast, random saturation and random hue. When predicting future actions, for those that exceed the length of the demonstration, we use a stationary action for padding. For the normalization of actions, we normalize them to the range of $[-1, 1]$. We also tried normalizing the actions to a Gaussian distribution $\mathcal{N}(0,1)$, but did not observe any performance improvement.

\section{Evaluation Details}
\subsection{Simulated Evaluation}
Evaluating robotic manipulation tasks is challenging due to the varying hardware, environments, and tasks used across different studies~\cite{li2024simpler}. While standardized real-world setups can be beneficial, they often require substantial time and resources~\cite{calli2015ycb, correll2016analysis, krotkov2018darpa, zhou2023train}. To address these issues while preserving the accuracy of real-world performance evaluations, SIMPLER~\cite{li2024simpler} has developed a system for evaluating manipulation policies trained on real data in simulated environments. Therefore, we employs SIMPLER for our evaluations, offering an easily reproducible and completely fair assessment framework.

\paravspace\paragraph{Task Definitions.}
We utilize all task variants provided in SIMPLER for our evaluations. The Google robot setup includes the following tasks: 1) ``pick Coke can'', 2) ``move \{obj1\} near \{obj2\}'', 3) ``(open / close) (top / middle / bottom) drawer'', and 4) ``open top drawer; place apple into top drawer''. The WidowX robot setup includes: 1) ``put the spoon on the towel'', 2) ``put carrot on plate'', 3) ``stack the green block on the yellow block'', and 4) ``put eggplant into yellow basket''. For the Google robot setup, both \textit{Visual Matching} (VM) and \textit{Variant Aggregations} (VA) evaluations are provided (See Sec.~4.1 in the main paper), while only the \textit{Visual Matching} (VM) evaluation is provided for the WidowX robot setup.

The total number of trials for each sub-task is shown in Table \ref{tab:num_trails in sim}. Note that to accommodate the limited number of the original trials (24 per task) on the WidowX robot, we repeat each original task trial five times (with different random initialization seeds) to enhance the statistical significance. For more detailed information on the tasks and evaluation protocols, we refer the readers to the SIMPLER paper~\cite{li2024simpler}. 

\begin{table}[!t]
\centering
\small
\caption{Simulated evaluation tasks and their number of trials in SIMPLER.}
\vspace{-4pt}
\begin{tabular}{l|cc}
\toprule
  Task &
   \multicolumn{2}{c}{\# Trials}\\ 
      &(VM) & (VA) \\ 
      \midrule
             Pick Coke Can     & \!\!\!\!\!\!300 &825     \\ 
             Move Near  & \!\!\!\!\!\!240 &600   \\ 
             Open/Close Drawer & \!\!\!\!\!\!216 &378   \\ 
             Open Top Drawer and Place Apple & \!\!\!\!\!\!108 &189   \\ 
             Put Spoon on Towel & ~~~~~120 {\footnotesize (24$\!\times\!$5)\!\!} & N/A   \\
             Put Carrot on Plate & ~~~~~120 {\footnotesize (24$\!\times\!$5)\!\!} &N/A   \\ 
             Stack Green Block on Yellow Block & ~~~~~120 {\footnotesize (24$\!\times\!$5)\!\!} &N/A   \\ 
             Put Eggplant in Yellow Basket & ~~~~~120 {\footnotesize (24$\!\times\!$5)\!\!} &N/A   \\ 
\bottomrule
\end{tabular}

\label{tab:num_trails in sim}
\end{table}

\paravspace\paragraph{Implementation Details.}
All simulated evaluations are conducted on a single NVIDIA A6000 GPU or a single NVIDIA A100 GPU. During inference, we employ DDIM~\cite{song2020denoising} sampling with 10 sampling steps and a classifier-free guidance (CFG)~\cite{ho2022classifier} coefficient of 1.5. 

\begin{figure}[!t]
    \centering
    \includegraphics[width=0.99\linewidth]{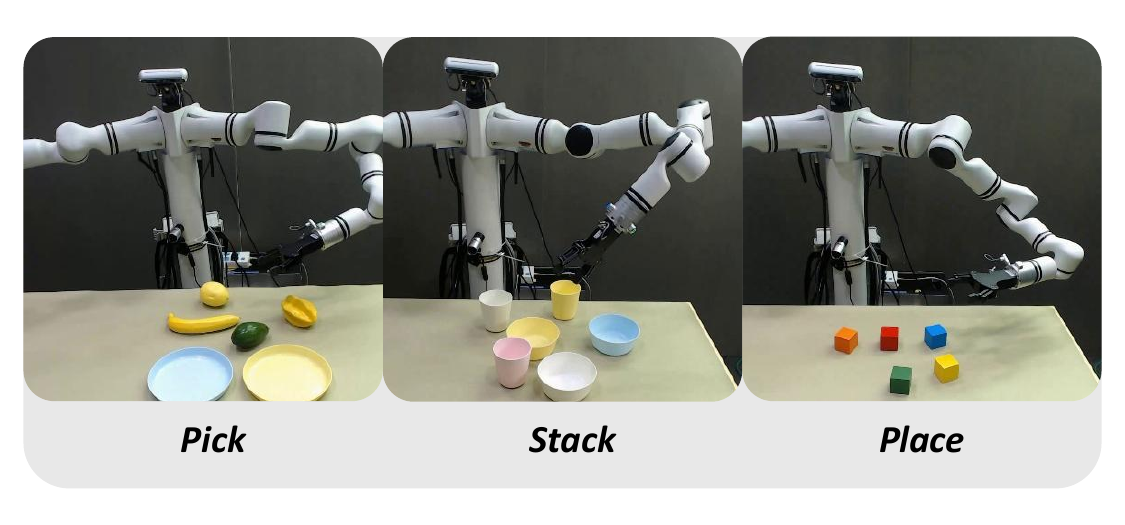}
    \vspace{-4mm}
    \caption{Seen task setups on the Realman robot.}
    \vspace{-2mm}
    \label{fig:seen_setting}
\end{figure}

\begin{figure}[!t]
    \centering
    \includegraphics[width=0.99\linewidth]{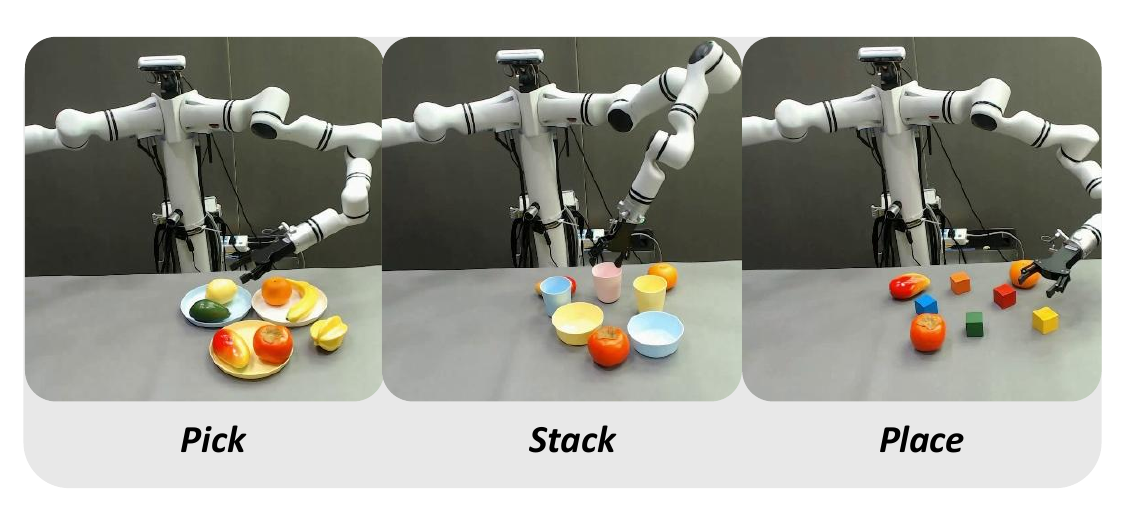}
    \vspace{-4mm}
    \caption{Unseen task setups on the Realman robot.}
    \vspace{-4mm}
    \label{fig:unseen_setting}
\end{figure}

We apply a unified adaptive action ensemble strategy (described in Sec.~3.4 in the main paper) across all our models, where the hyperparameter $\alpha$ is set to $0.1$. The ensemble window size $K$ determines the number of historical observations and their predicted actions to be used. Since different dataset have varying control frequency and robot speed, the window size $K$ should be adaptive. We select the window size $K$ such that the product of $K$ and the average standard deviation ($std$) of the 6D action per timestep remains constant across different datasets. Formally, this relationship can be expressed as $C=K\times std$, where $C$ is a constant representing the distance and angle traversed by the robot over the last $K$ steps. Empirically, we set the $C$ to $0.2$ for all experiments and derive their $K$ accordingly.

For the experiments on the Google robot, the task success rates of the pervious methods were   evaluated by \cite{li2024simpler} and incorporated into this paper, except for OpenVLA ~\cite{kim2024openvla}, which was not included in~\cite{li2024simpler}. For OpenVLA, we directly evaluate it using the weight from its official repository\footnote{\href{https://huggingface.co/openvla/openvla-7b-prismatic}{https://huggingface.co/openvla/openvla-7b-prismatic}}. On the WidowX robot, the performance of RT-1-X ~\cite{o2023open_x_embodiment} is directly reported as in \cite{li2024simpler} and OpenVLA is evaluated as described above. For Octo-Base and Octo-Small~\cite{team2024octo}, we incorporate more random seeds and conduct five retests to mitigate volatility in their probabilistic sampling. For all configurations of our models, we evaluate them using the same number of tests as those used for Octo. Note that the performance of RT-2-X on WidowX has not been reported~\cite{li2024simpler}, and no model weights are publicly available for evaluation.

\begin{figure*}[!t]
    \centering
    \includegraphics[width=0.99\linewidth]{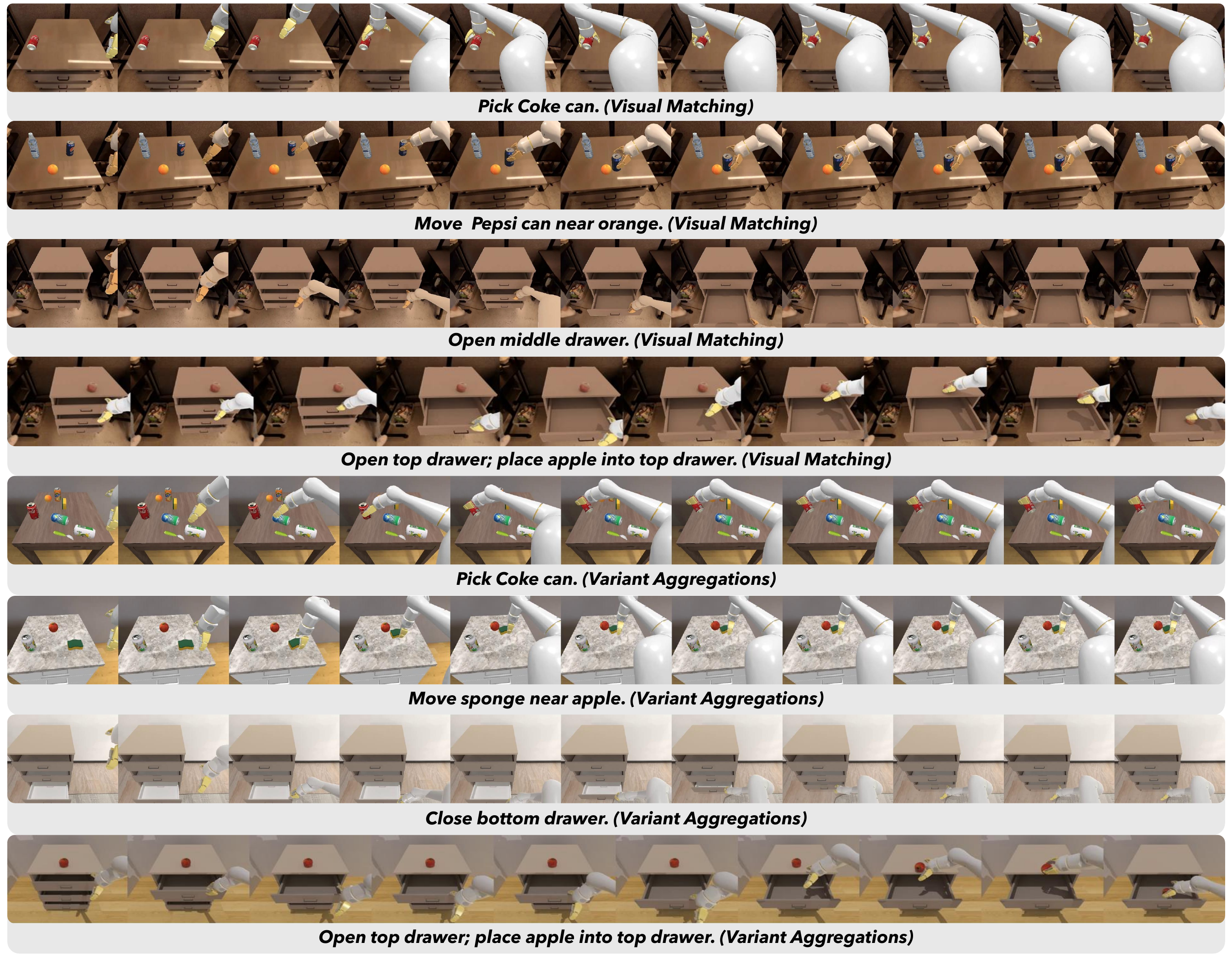}
    \vspace{-2mm}
    \caption{Visual examples of each task on the Google robot driven by our model.}
    \label{fig:google_robot_visual_video}
\end{figure*}

\begin{figure*}[!t]
    \centering
    \includegraphics[width=0.99\linewidth]{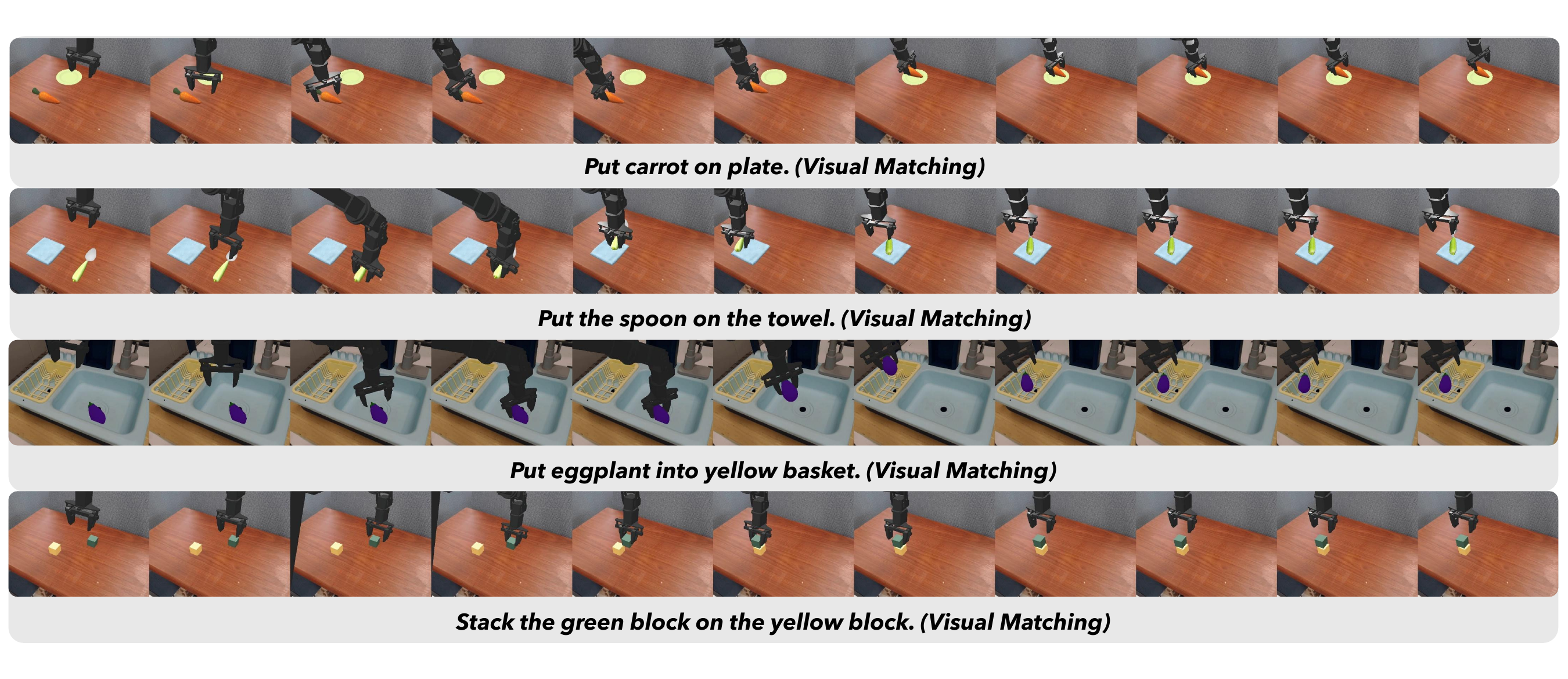}
    \vspace{-2mm}
    \caption{Visual examples of each task on the WidowX robot driven by our model.}
    \label{fig:widowx_visual_video}
\end{figure*}

\paravspace\paragraph{Visualization Results.} Figure \ref{fig:google_robot_visual_video} provides the successful examples 
 of each task executed by our default model on the Google robot, while Figure \ref{fig:widowx_visual_video} presents those on the WidowX robot.

\subsection{Real-World Evaluation}
\begin{figure*}[!t]
    \centering
    \includegraphics[width=0.99\linewidth]{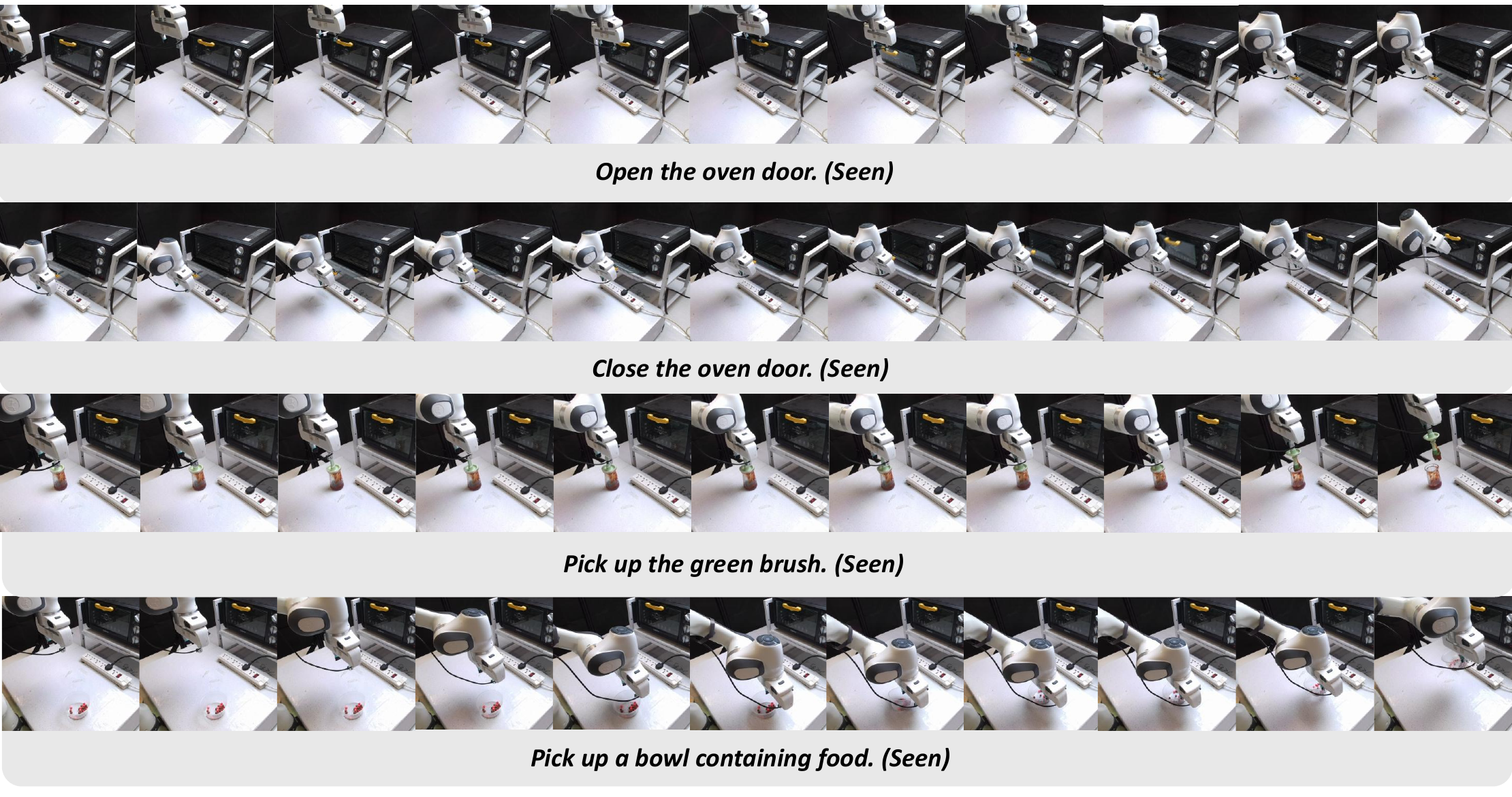}
    \caption{Visual examples of real-world tasks on the Franka robot using our model.} 
    \label{fig:franka_eval}
\end{figure*}

\paragraph{Task Definitions.}
Section~4.2 in the main paper defines three tasks ``Pick'', ``Stack'' and ``Place'' on the Realman robot. The setups of these tasks in the seen environment are illustrated in Figure \ref{fig:seen_setting}, while those on the unseen table are illustrated in Figure~\ref{fig:unseen_setting}. Section~4.3 in the main paper defines four tasks:
``Close the oven door'', ``Open the oven door'', ``Pick up the green brush'' and ``Pick up a bowl containing food''. The setups of these tasks are presented in Figure \ref{fig:franka_eval}.

\paravspace\paragraph{Implementation Details.}
The finetuning process was conducted on 16 NVIDIA A100 GPUs, with all models involved in the comparison utilizing PyTorch FSDP for full finetuning and a batch size of 256. For our method, we simply tested one finetuned checkpoint at 10K finetuning steps, which takes 7.5 hour's finetuning time. For Octo~\cite{team2024octo} and OpenVLA~\cite{kim2024openvla}, we tested multiple finetuned checkpoints at different steps and select the ones that worked best in the real-world tasks. Specifically, for the Realman robot, Octo uses the 20K checkpoint, while OpenVLA uses the 30K checkpoint; both models use the 30K checkpoint on the Franka robot. All finetuning data follows the same data augmentation as used during pretraining.

\begin{figure*}[!t]
    \centering
    \includegraphics[width=0.99\linewidth]{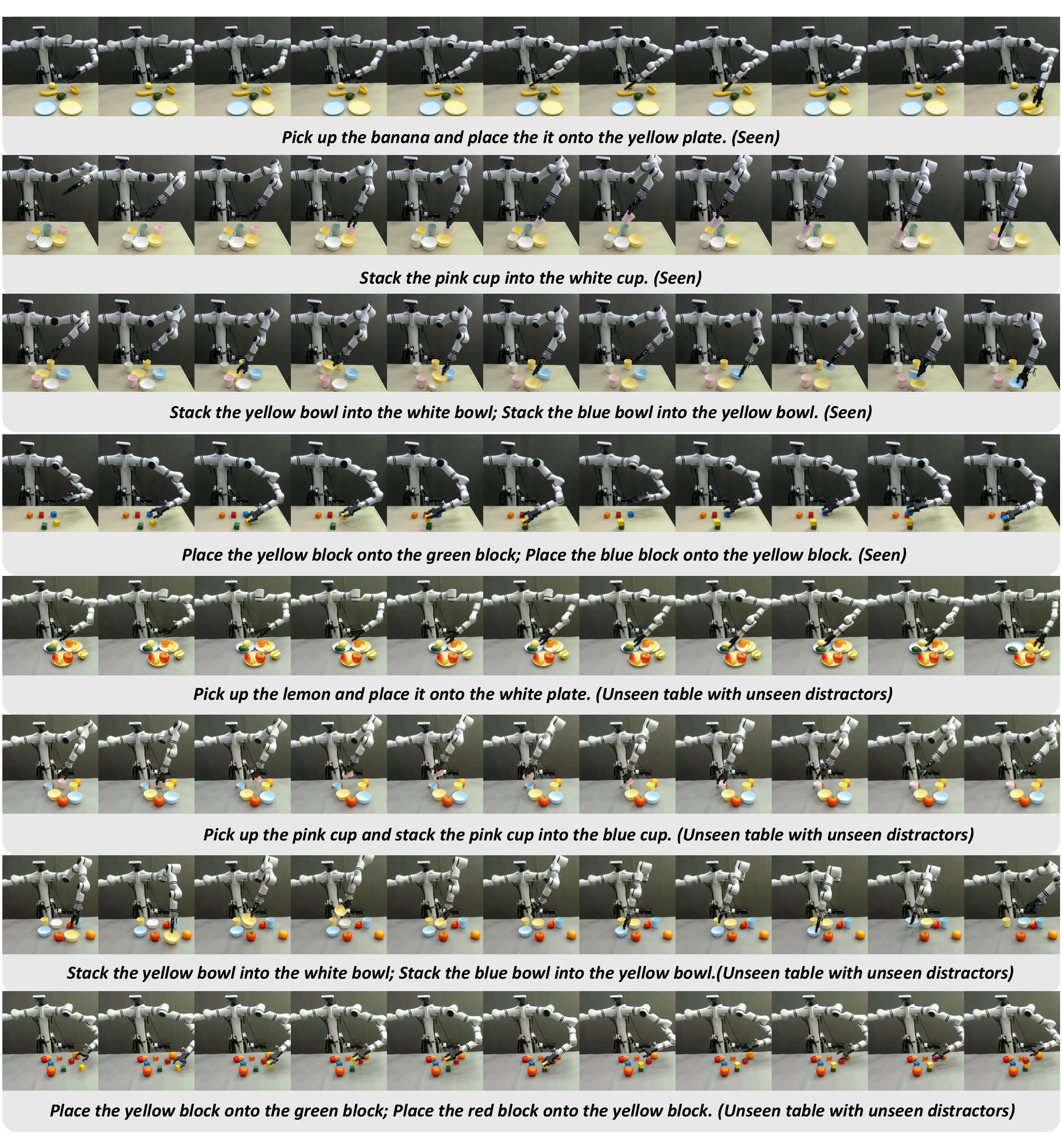}
    \caption{Examples of real-world seen tasks and unseen-table-with-unseen-distractors tasks on the Realman robot driven by our model.}
    \label{fig:realman_eval}
\end{figure*}

\begin{figure*}[!t]
    \centering
    \includegraphics[width=0.99\linewidth]{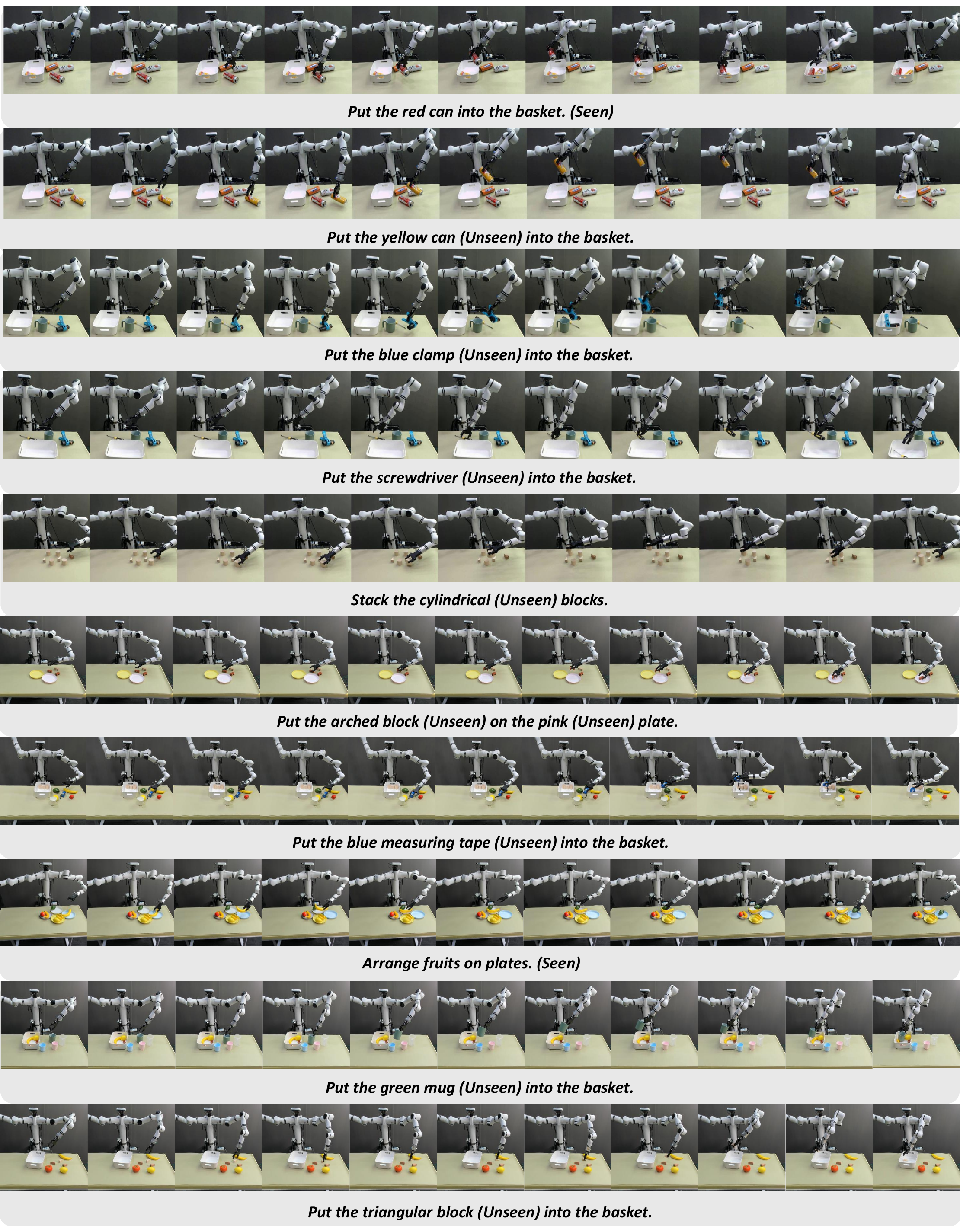}
    \caption{More examples of different real-world tasks on the Realman robot using our model.}
    \label{fig:realman_more_visual}
\end{figure*}

\paravspace\paragraph{Visualization results.} Figure~\ref{fig:realman_eval} and~\ref{fig:realman_more_visual} present evaluation examples of each tasks executed by our default model on the Realman robot, while Figure~\ref{fig:franka_eval} provides evaluation examples of each tasks on the Franka robot.

 \begin{table}[!t]
\centering
\footnotesize  
\caption{The hyperparameters of different action modules.
}
\vspace{-4pt}
\begin{tabular}{lccccc}
\toprule
  Action Model  &
   \# Layers&
   Emb Dim  & \# Heads & \# Params\\ \midrule
             MLP (3-Layer)     & 3     & 256    & N/A     & 3M      \\ 
             MLP (7-Layer)     & 7     & 1024    & N/A     & 89M      \\ 
             DiT-Small     & 6     & 384    & 4     & 13M     \\ 
             DiT-Base     & 12     & 768    & 12     & 89M      \\ 
             DiT-Large  & 24  &1024  &16   &  308M   \\ 
\bottomrule
\end{tabular}
\label{tab:action_model_hyperparam}
\end{table}

\section{Ablation Study Details}
\subsection{Action Model Architectures}

As described in Sec.~4.4 in the main paper, we study the performances of different action models on the Google robot and WidowX robot. Table \ref{tab:action_model_hyperparam} presents the hyperparameters of all the action modules mentioned in Sec.~4.4. The MLP components of these models are structured such that each MLP block expands the embedding dimension by four times and then scales it back to the original embedding dimension. The structure of the MLP (3-layer) model is consistent with the model architecture in Octo~\cite{team2024octo}. Table~7 in the main paper illustrates that such small action models have significant limitations in modeling actions and perform far worse than the DiT models.

\begin{table}[!t]
\centering
\small
\caption{Comparison of different classifier-free guidance coefficients ~\cite{ho2022classifier} in simulated evaluations.
}
\vspace{-4pt}
\begin{tabular}{l|ccc|c}
\toprule
   \multirow{2}{*}{CFG Scale}&
   \multicolumn{2}{c}{GR}   &
   WR   &  \multirow{2}{*}{Average}\\
      &(VM) & (VA)& (VM)&  \\ 
      \midrule
             1.0     & 66.9     & 54.0     & 46.3     & 55.7     \\ 
             1.5  & 74.8   & 61.3 & \textbf{51.3} & 62.5   \\ 
              3.0     &\textbf{76.6}  &\textbf{63.1}     & 48.3      &\textbf{62.7}   \\
                
\bottomrule
\end{tabular}
\vspace{-8pt}
\label{tab:cfg_ablation}
\end{table}

\subsection{Classifier-Free Guidance Scale}
During inference, the classifier-free guidance (CFG) is always applied with a default scale of $1.5$, as the experiments show that incorporating the CFG technique significantly improves success rates. Different CFG scales are evaluated in Table~\ref{tab:cfg_ablation}.

\clearpage



\end{document}